\documentclass[11pt]{article}
\usepackage[preprint]{acl}
\usepackage{times}
\usepackage{latexsym}
\usepackage[T1]{fontenc}
\usepackage[utf8]{inputenc}
\usepackage{microtype}
\usepackage{inconsolata}
\usepackage{amsmath,amssymb,mathtools}
\usepackage{booktabs}
\usepackage{multirow}
\usepackage{graphicx}
\usepackage{algorithm}
\usepackage{algpseudocode}
\usepackage{array}
\usepackage{makecell}
\usepackage{xurl}
\title{When to Review: Spaced Repetition for Continual Pre-Training of Language Models}
\author{Alankar Atreya\textsuperscript{1} \quad
  Devesh Batra\textsuperscript{1} \quad
  Yoages Kumar Mantri\textsuperscript{1} \quad
  Geremy Bantug\textsuperscript{1} \\
  \bfseries Greig A Cowan\textsuperscript{1} \quad
  Raad Khraishi\textsuperscript{1,2} \\[0.8em]
  \normalfont\textsuperscript{1}NatWest AI Research \quad
  \textsuperscript{2}University College London}
\newcommand{\dold}{\mathcal{D}_{\mathrm{old}}}
\newcommand{\dnew}{\mathcal{D}_{\mathrm{new}}}

\newcommand{\batch}{\mathcal{B}}
\newcommand{\due}{\mathcal{M}}
\newcommand{\srt}{\textsc{SRT}}
\newcommand{\cpt}{\textsc{CPT}}
\newcommand{\base}{\textsc{Base}}

\newcommand{\uniform}{\textsc{Uniform Replay}}
\newcommand{\ppl}{\mathrm{PPL}}
\newcommand{\risk}{\mathcal{R}}
\newcommand{\old}{\mathrm{old}}
\newcommand{\new}{\mathrm{new}}
\newcommand{\bftab}[1]{\textbf{#1}}

\begin{document}
\maketitle
\begin{abstract}
Continual pre-training of large language models must acquire new information without erasing old knowledge. Existing replay methods often choose a global old/new mixture and sample uniformly, ignoring that examples differ in how quickly they are forgotten. We formulate continual pre-training as adaptive review scheduling: the training loop should decide not only how much history to replay, but which examples should return at each step. We introduce Spaced Repetition Training (SRT), a continual learning framework inspired by cognitive science, which schedules sample-rehearsal using the SuperMemo-2 (SM-2) algorithm. SRT maintains per-example review state, maps per-example perplexity to a recall-quality signal, and schedules historical examples for retention and new examples for consolidation while leaving the model, objective, and optimizer unchanged. On temporally separated Wikipedia and code corpora, SRT improves the stability-plasticity trade-off, recovering 5 to 37 percentage points of old-knowledge accuracy lost by naive continual pre-training across model scales while preserving or improving new-knowledge acquisition. At larger scale, SRT preserves broad benchmark performance that naive continual pre-training and uniform replay substantially degrade. Experiments with vision and tabular data further suggest that the scheduling principle extends beyond language when paired with an appropriate recall signal.
\end{abstract}

\section{Introduction}
\label{sec:introduction}

Large language models are increasingly updated after pre-training as facts change, software ecosystems evolve, and new domain corpora appear \citep{gururangan2020dont, ibrahim2024simple}. Continual pre-training (CPT) is cheaper than training from scratch, but updates on new data can erase previously learned knowledge: catastrophic forgetting \citep{mccloskey1989catastrophic, kirkpatrick2017overcoming}. In LLMs, this forgetting can also degrade broad capabilities such as multitask reasoning and factual recall \citep{li2024examining, ibrahim2024simple}.

Replay is an attractive mitigation strategy because it is architecture-agnostic and scales naturally to LLM training \citep{rolnick2019experience, chen2025effective, li2025tic}. Most replay methods, however, operate at the mixture level: they choose how much historical data to include and then sample old examples uniformly or by a fixed curriculum. This ignores heterogeneity in retention difficulty. Some examples remain stable after one review, while others drift quickly under interference from new learning. Uniform replay can therefore spend budget on already-retained examples while missing fragile ones.

We instead treat CPT as an adaptive review-scheduling problem. Drawing on spaced repetition and difficulty-adaptive memory consolidation \citep{ebbinghaus1885memory, cepeda2006distributed}, we ask which examples should be reviewed at the current training step, not only how much old data should be mixed in. Algorithms such as SuperMemo-2 already operationalize this idea for human memory through per-item review state \citep{wozniak1998supermemo}; we adapt the same principle to language model training.

We introduce \textbf{Spaced Repetition Training} (SRT), a drop-in scheduler for continual pre-training. Each replayable example maintains an ease factor, review count, interval, and due step. At review time, per-example perplexity is converted into a recall-quality score that updates the next interval: difficult examples return sooner, while confidently retained examples receive longer gaps. The scheduler applies symmetrically to historical examples for retention and new examples for consolidation, and it leaves the model architecture, objective, and optimizer unchanged.

\newpage
\paragraph{Our contributions are as follows:}
\begin{itemize}
    \item We formulate replay-based LLM continual pre-training as adaptive review scheduling, motivated by spaced repetition and heterogeneous retention difficulty.
    \item We propose SRT, a sample-level SuperMemo-2 scheduler that converts per-example perplexity into recall quality and schedules both old and new examples without changing the model, loss, or optimizer.
    \item We show that SRT mitigates catastrophic forgetting more effectively than naive CPT and uniform replay while maintaining new-knowledge acquisition on temporally separated Wikipedia and code corpora, and that it preserves broad benchmark performance that both baselines degrade. Additional vision and tabular experiments suggest that the scheduling principle generalises beyond language.
\end{itemize}

\section{Related Work}
\label{sec:related-work}

\paragraph{Catastrophic forgetting.} Sequential training can overwrite representations needed for earlier tasks \citep{mccloskey1989catastrophic}. Standard mitigation strategies include regularization, which protects important parameters \citep{kirkpatrick2017overcoming, zenke2017continual}; architectural isolation or expansion \citep{rusu2016progressive, hung2019compacting}; and replay, which interleaves previous examples with new data \citep{rolnick2019experience, parisi2019continual}. For LLM CPT, replay is especially practical because task boundaries are weak, parameter growth is undesirable, and gradient-level regularization is expensive. SRT remains within the replay family but shifts the design question from mixture size to per-example review timing.

\paragraph{Continual pre-training of LLMs.}
Domain-adaptive pre-training can improve target-domain performance while shifting models away from prior capabilities \citep{gururangan2020dont}. Recent CPT work studies learning-rate rewarming and re-decay, data mixtures, replay ratios, curricula, and temporal benchmarks \citep{ibrahim2024simple, chen2025effective, li2025tic}. Other studies show that continual updates can degrade established benchmarks such as multitask language understanding and factual recall \citep{li2024examining}. These methods primarily tune optimization or mixture-level replay. SRT is complementary: given a replay budget, it decides which stored examples should be reviewed now.

\paragraph{Human-inspired replay.} SuperMemo-2 maps recall quality to expanding inter-review intervals through per-item state \citep{wozniak1998supermemo}. Recent neural methods import related ideas via self-synthesized rehearsal, Leitner-style queues, or forgetting-curve replay \citep{huang2024mitigating, mhamdi2024leitner, feng2026forever}. SRT differs by maintaining SuperMemo-2 state for individual training examples and deriving recall quality directly from per-example perplexity computed in the standard forward pass, requiring no synthesis pipeline or held-out probe set.

\section{Method}
\label{sec:method}

\subsection{Problem Setup}
\label{sec:problem-setup}

Let $p_{\theta}$ denote a language model with parameters $\theta$ initialized from a pre-trained checkpoint. We consider a continual pre-training setting in which the model is updated on an incoming corpus $\dnew = \{x_j^{\new}\}$ while also retaining access to a historical corpus $\dold = \{x_i^{\old}\}$ representing knowledge acquired before the update. The goal of continual pre-training is not simply to minimize next-token loss on $\dnew$, but to acquire new information while preserving performance on $\dold$. We express this as a weighted combination of evaluation risks,
\begin{equation}
\min_{\theta'} \; \lambda \, \risk_{\old}(\theta') + (1 - \lambda) \, \risk_{\new}(\theta'),
\label{eq:risk}
\end{equation}
where $\risk_{\old}$ and $\risk_{\new}$ are evaluation risks on old and new data respectively, and $\lambda \in [0, 1]$ controls the stability-plasticity trade-off.

In practice, training operates under a finite budget of optimization steps and tokens. At step $t$, a mini-batch $\batch_t$ of size $B$ must allocate exposure between old and new examples,
\begin{equation}
\batch_t = \batch^{\old}_t \cup \batch^{\new}_t, \qquad |\batch^{\old}_t| + |\batch^{\new}_t| \leq B.
\end{equation}
Naive continual pre-training sets $|\batch^{\old}_t| = 0$ and spends the entire batch on $\dnew$. Replay-based methods instead allocate a fraction of each batch to examples drawn from $\dold$. The central design question is therefore not only how much of each batch to allocate to old data, but which specific examples should be drawn from each pool at step $t$, given that retention difficulty varies across both corpora.

\subsection{Spaced Repetition Training}
\label{sec:srt-algorithm}

SRT addresses this question by maintaining review state and scheduling using the SuperMemo-2 algorithm \citep{wozniak1998supermemo}. Each example $x_i \in \dold \cup \dnew$ is associated with a review state
\begin{equation}
s_i(t) = \bigl(E_i(t), n_i(t), I_i(t), d_i(t)\bigr),
\end{equation}
where $E_i$ is the ease factor controlling interval growth, $n_i$ is the number of consecutive successful reviews, $I_i$ is the current inter-review interval measured in training steps, and $d_i$ is the next due step. Following standard SuperMemo-2 initialization, we set $E_i = 2.5$, $n_i = 0$, and $I_i = 1$, with initial due times staggered uniformly over the first few training steps to avoid reviewing all items simultaneously. State is maintained symmetrically for examples in $\dold$ and $\dnew$, so SRT applies the same scheduling logic to historical examples (where it determines replay timing) and to incoming examples (where it determines consolidation timing).

\paragraph{Due-set sampling.}
At training step $t$, we define the sets of examples in $\dold$ and $\dnew$ whose next review is due,
\begin{align}
\due_t^{\old} &= \{x_i \in \dold : d_i \leq t\}, \\
\due_t^{\new} &= \{x_j \in \dnew : d_j \leq t\}.
\end{align}
The training batch at step $t$ allocates $\lfloor \rho B \rfloor$ slots to due old examples sampled from $\due_t^{\old}$ and the remaining slots to due new examples sampled from $\due_t^{\new}$, where $\rho \in [0, 1]$ is the target old-exposure fraction. Both old and new examples are therefore selected by the scheduler, not only those drawn from $\dold$. If either due set is empty at step $t$, the unused budget is released to the other stream, ensuring that scheduling never blocks training progress.

\paragraph{Perplexity-derived recall quality.}
When a selected example $x_i = (x_{i,1}, \ldots, x_{i,T_i})$ is reviewed, we compute its token-average negative log-likelihood under the current model,
\begin{equation}
L_i(t) = -\frac{1}{T_i} \sum_{k=1}^{T_i} \log p_{\theta_t}(x_{i,k} \mid x_{i,<k}),
\label{eq:nll}
\end{equation}
and its perplexity $\ppl_i(t) = \exp(L_i(t))$. Perplexity is computed during the standard forward pass and requires no auxiliary evaluation. We convert perplexity into the discrete recall quality score $q_i(t) \in \{1, 2, 3, 4, 5\}$ used by SuperMemo-2 via monotonically increasing thresholds $0 < \tau_5 < \tau_4 < \tau_3 < \tau_2 < \tau_1$ that partition model performance into quality levels,
\begin{equation}
q_i(t) = \max\!\left(0, \; 5 - \sum_{k=1}^{5} \mathbf{1}\!\left[\ppl_i(t) \geq \tau_k\right]\right).
\label{eq:pplscore}
\end{equation}
Lower perplexity yields higher recall quality, with $q = 5$ corresponding to fluent retention and $q = 0$ to essentially no model on the example. Threshold values were selected empirically; Section~\ref{sec:ablations} analyses sensitivity to their scaling.

The recall quality $q_i(t)$ is computed from the training-time loss at the current parameters $\theta_t$, that is, from the same forward pass used to obtain the gradient, before the optimizer step that produces $\theta_{t+1}$. It therefore reflects how well the model retained $x_i$ prior to the current update, not how well it fits $x_i$ after being trained on it.

For non-generative models where perplexity is not defined, the same scheduler can be applied by replacing $\mathrm{PPL}_i(t)$ with an alternative confidence-based signal. Details of this adaptation are provided in Appendix~\ref{app:vision-confidence}.

\paragraph{Review state update.}
Given recall quality $q_i$, the ease factor is updated according to the SuperMemo-2 rule,
\begin{align}
\Delta(q_i) &= 0.1 - (5 - q_i)\bigl(0.08 + 0.02(5 - q_i)\bigr), \label{eq:delta} \\
E_i &\leftarrow \max\bigl(1.3, \, E_i + \Delta(q_i)\bigr). \label{eq:ease}
\end{align}
If $q_i < 3$, the example is treated as forgotten and we reset $n_i \leftarrow 0$ and $I_i \leftarrow 1$, scheduling immediate re-review. Otherwise we increment $n_i$ and update the interval according to
\begin{equation}
I_i \leftarrow \begin{cases}
1, & n_i = 1, \\
6, & n_i = 2, \\
\lceil I_i E_i \rceil, & n_i > 2.
\end{cases}
\label{eq:interval}
\end{equation}
The next review is scheduled by setting $d_i \leftarrow t + I_i$. The model parameters $\theta$ are updated using standard next-token cross-entropy loss on $\batch_t$, identical to the underlying language modelling objective. SRT therefore introduces no additional loss terms, no architectural changes, and no learned parameters beyond those already present in $p_\theta$. The full procedure is summarized in Algorithm~\ref{alg:srt}.

\begin{algorithm}[t]
\small
\caption{One SRT training step.}
\label{alg:srt}
\begin{algorithmic}[1]
\Require corpora $\dold$, $\dnew$, old-exposure cap $\rho$, batch size $B$, step $t$
\State $\due_t^{\old} \leftarrow \{x_i \in \dold : d_i \leq t\}$ \Comment{due old examples}
\State $\due_t^{\new} \leftarrow \{x_j \in \dnew : d_j \leq t\}$ \Comment{due new examples}
\State $b^{\old} \leftarrow \min\bigl(|\due_t^{\old}|, \, \lfloor \rho B \rfloor\bigr)$
\State sample $\batch_t^{\old} \subseteq \due_t^{\old}$ with $|\batch_t^{\old}| = b^{\old}$
\State sample $\batch_t^{\new} \subseteq \due_t^{\new}$ to fill remaining slots
\State $\batch_t \leftarrow \batch_t^{\old} \cup \batch_t^{\new}$
\For{$x_i \in \batch_t$} \Comment{scored at $\theta_t$, before the update}
    \State compute $\ppl_i(t)$ via Eq.~\ref{eq:nll} and $q_i(t)$ via Eq.~\ref{eq:pplscore}
\EndFor
\State update $\theta$ on $\batch_t$ using next-token loss \Comment{$\theta_t \rightarrow \theta_{t+1}$}
\For{$x_i \in \batch_t$} \Comment{update review state only}
    \State update $E_i$, $n_i$, $I_i$ via Eqs.~\ref{eq:ease}--\ref{eq:interval}
    \State set $d_i \leftarrow t + I_i$
\EndFor
\end{algorithmic}
\end{algorithm}

\section{Experimental Setup}
\label{sec:setup}

\subsection{Corpora}
\label{sec:corpora}

We construct temporally grounded corpora in two domains where knowledge evolves measurably over time: encyclopedic text and source code. Both domains are split into an \emph{old} corpus $\dold$ representing information available before a temporal cutoff and a \emph{new} corpus $\dnew$ representing post-cutoff information, enabling evaluation that distinguishes retained knowledge from acquired knowledge.

\paragraph{Wikipedia.}
For encyclopedic text, we align timestamped English Wikipedia snapshots by normalized article title, producing aligned old-new pairs corresponding to the same entity at different points in time. This formulation follows temporal knowledge benchmarks that evaluate whether language models can acquire updated information while retaining prior knowledge \citep{jang2022temporalwiki, cheng2024dated, li2025tic}. The new-data corpus is extracted from sentence-level diffs between aligned pairs, capturing inserted spans and the updated side of modified spans, which preserves localized factual updates while filtering unchanged text. Full snapshot selection, topic stratification, and diff-extraction details are provided in Appendix~\ref{app:wikipedia-corpora}.

\paragraph{Code.}
For source code, we construct temporally separated corpora from GitHub repositories selected by creation and commit timestamps. Old examples are drawn from repositories that existed before January~2022, and new examples from repositories created after July~2024, with verification that $\dnew$ repositories did not exist before the cutoff window. Repositories span thirteen widely used programming languages. Unlike the Wikipedia corpus, old and new code examples are not aligned revisions of the same artifact; they represent temporally distinct samples from evolving software distributions, reflecting how software ecosystems shift through new libraries, repositories, and conventions rather than incremental edits to existing artifacts. Full repository selection criteria and language coverage are provided in Appendix~\ref{app:code-corpora}.

\paragraph{Corpus statistics.}
Corpus sizes for both domains are summarized in Appendix~\ref{app:wikipedia} (Table~\ref{tab:corpora-summary}). For Wikipedia, the old and new training corpora each contain $31{,}729$ aligned examples. For code, the old corpus contains $22{,}055$ examples and the new corpus contains $70{,}643$ examples.

\subsection{Models and Training}
\label{sec:models-and-training}

We evaluate SRT across two model scales. Smaller-scale experiments use TinyLlama-1.1B-Chat \citep{zhang2024tinyllama}, which provides a fast iteration platform for ablations and main comparisons. Larger-scale experiments use Llama-3.2-3B-Instruct \citep{grattafiori2024llama, meta2024llama32}, a 3-billion-parameter instruction-tuned model released by Meta with a knowledge cutoff of December~2023. We continually pre-train from the instruction-tuned checkpoint rather than a base checkpoint because updating already-deployed instruct models on fresh data is the realistic practitioner scenario our method targets.

All conditions use the same continual pre-training recipe, following established practice for replay-based LLM continual pre-training \citep{ibrahim2024simple}. We use causal language modeling with the next-token cross-entropy objective, AdamW optimization, linear learning-rate scheduling with warmup and decay, gradient clipping, and bfloat16 precision where supported. Hyperparameters were tuned on a held-out development split and held fixed across all baselines and SRT conditions to ensure that performance differences reflect the replay strategy rather than optimization differences. Complete hyperparameter values are provided in Appendix~\ref{app:training-details}.

\subsection{Baselines}
\label{sec:baselines}

We compare four conditions against the original pre-trained checkpoint. \textbf{Base} is the pre-trained model before continual update. It establishes the starting point for both old-knowledge retention and new-knowledge acquisition. \textbf{CPT} is naive continual pre-training, where each training batch is drawn entirely from $\dnew$. This represents the standard no-replay baseline and exhibits the canonical forgetting failure mode. \textbf{Uniform Replay} draws each batch using a nominal 20/80 old/new exposure ratio, with old examples sampled uniformly at random from $\dold$ and new examples sampled uniformly at random from $\dnew$. This baseline shares the same nominal replay budget and pool structure as SRT but uses no per-example scheduling for either pool, isolating the contribution of adaptive review timing. \textbf{PPL-Prioritised} is a difficulty-aware replay baseline that isolates whether SRT's gains come from spaced-repetition scheduling or simply from attending to hard examples. It uses the same 20/80 old/new exposure ratio as Uniform Replay and SRT, but within each pool it selects the highest-perplexity examples rather than sampling uniformly, without maintaining any SM-2 review state or interval scheduling. It therefore replays the currently hardest examples at every step. Because its role is to separate difficulty-prioritisation from interval-based scheduling rather than to serve as a scale-general baseline, we evaluate it on both TinyLlama settings (Wikipedia and code) as a diagnostic control. \textbf{SRT} uses the same nominal 20/80 old/new exposure cap and pool structure as Uniform Replay, but selects examples in each batch using SuperMemo-2 due times and perplexity-derived recall quality as described in Section~\ref{sec:srt-algorithm}. Both old and new examples are selected by the scheduler. The old-exposure fraction is set to $\rho = 0.2$, matching Uniform Replay so that the only difference between the two conditions is whether within-pool selection is scheduled or uniform.

\subsection{Evaluation}
\label{sec:evaluation}

We evaluate along two complementary axes. Source-grounded question answering measures retention and acquisition on the temporally split corpora. Standard capability benchmarks measure whether continual updates degrade broad model abilities beyond the updated domains.

\paragraph{Source-grounded QA.}
For each of the Wikipedia and code domains, we generate multiple-choice QA benchmarks in MMLU-style format directly from source passages in $\dold$ and $\dnew$, producing matched old and new evaluation splits of $500$ questions each. Questions are generated from source content and manually checked for answerability and label support. We report accuracy as the mean across questions together with the bootstrap standard deviation computed from $10{,}000$ resamples.

\paragraph{Broad capability benchmarks.}
We evaluate on four widely used benchmarks: MMLU \citep{hendrycks2021mmlu} for multitask knowledge and reasoning, BBH \citep{suzgun2023challenging} for hard reasoning tasks, GSM8K \citep{cobbe2021training} for grade-school mathematical reasoning, and PIQA \citep{bisk2020piqa} for physical commonsense. All broad capability benchmarks are evaluated using the EleutherAI LM Evaluation Harness \citep{eval-harness}, which provides standardized few-shot prompts, scoring procedures, and task implementations across models. Using a single evaluation framework ensures that benchmark scores are directly comparable across baselines and SRT, and that our reported numbers are reproducible by other researchers. These benchmarks are not optimization targets for any condition and serve to detect whether continual updates damage capabilities unrelated to the updated domains.

\section{Results}
\label{sec:results}

We evaluate SRT along two axes: source-grounded temporal QA (Section~\ref{sec:temporal-qa-results}) and broad capability benchmarks (Section~\ref{sec:broad-results}).

\subsection{Temporal QA: Retention and Acquisition}
\label{sec:temporal-qa-results}

Table~\ref{tab:temporal-qa} reports source-grounded QA accuracy across both models. The pattern is consistent at both scales: naive CPT improves new-knowledge accuracy at the cost of retention, uniform replay partially mitigates the drop, and \srt{} recovers retention while preserving or improving new-knowledge accuracy.

\paragraph{Naive CPT degrades retention.}
On TinyLlama, \cpt{} raises new accuracy from $13.1\%$ to $17.0\%$ on Wikipedia data but old accuracy collapses from $54.3\%$ to $11.7\%$. The same pattern holds on Llama-3.2-3B-Instruct, where old Wikipedia accuracy drops from $50.2\%$ to $43.0\%$.

\paragraph{Scheduling, not exposure, drives SRT's gains.}
\uniform{} reaches $25.2\%$ and $46.2\%$ old Wikipedia accuracy on TinyLlama and Llama-3.2-3B-Instruct respectively, in both cases substantially above \cpt{} but below \srt{} ($49.0\%$ and $51.6\%$). Because \uniform{} and \srt{} share the same $20/80$ exposure cap and pool structure and differ only in within-pool selection, this gap is attributable to scheduling rather than exposure. The gap is larger at the smaller model scale ($23.8$ percentage points on TinyLlama versus $5.4$ on Llama-3.2-3B-Instruct), suggesting that smaller models benefit more from prioritized review while larger models retain more knowledge under uniform replay alone.

\paragraph{Scheduling versus difficulty alone.}
To separate interval-based scheduling from simple difficulty-prioritisation, we compare \srt{} against a PPL-Prioritised baseline that replays the highest-perplexity old examples without SM-2 review state (Table~\ref{tab:temporal-qa}). Difficulty-awareness alone is clearly beneficial: on Wikipedia, PPL-Prioritised improves old-knowledge retention over \uniform{} ($27.4\%$ vs.\ $25.2\%$) and achieves the highest new-knowledge accuracy of any method ($38.8\%$). However, the two difficulty-aware methods occupy different points on the stability-plasticity spectrum. PPL-Prioritised is markedly more plastic, acquiring new knowledge aggressively but recovering far less old knowledge than \srt{} ($27.4\%$ vs.\ $49.0\%$ on Wikipedia, a $21.6$-point retention gap). We attribute this to its greedy focus on the currently hardest examples, which neglects intermediate-difficulty examples that \srt{}'s interval scheduling continues to revisit along their forgetting trajectory. On code, \srt{} outperforms PPL-Prioritised on all three metrics. Overall, \srt{} achieves the best old-knowledge retention on both domains and the best combined score, while PPL-Prioritised attains stronger new-knowledge accuracy on Wikipedia; interval-based scheduling thus provides a substantially better retention-acquisition balance than one-shot difficulty ranking, though the two methods prioritise that balance differently.

\paragraph{SRT improves both retention and acquisition.}
\srt{} produces the best combined score on every model-domain pair in Table~\ref{tab:temporal-qa}. On Wikipedia, it restores old accuracy to or above the base-model level at both scales while matching or exceeding \cpt{} on new accuracy. On code QA, \srt{} is the only method that simultaneously improves both old and new accuracy relative to \cpt{} on both models. \uniform{} is competitive with \cpt{} on code at both scales but trails \srt{} consistently, confirming that scheduling provides value across model scales even when uniform replay alone is adequate. To corroborate these findings on a benchmark we did not construct, we additionally evaluate on TemporalWiki \citep{jang2022temporalwiki}, an external factual-probe set. \srt{} substantially limits the perplexity degradation that \cpt{} incurs on this benchmark, consistent with the retention advantage measured on our source-grounded QA. Full results and caveats are provided in Appendix~\ref{app:temporalwiki}.

\begin{table*}[t]
\centering
\small
\setlength{\tabcolsep}{3.5pt}
\begin{tabular}{llrrrrrr}
\toprule
& & \multicolumn{3}{c}{Wikipedia QA} & \multicolumn{3}{c}{Code QA} \\
\cmidrule(lr){3-5}\cmidrule(lr){6-8}
Model & Method & Old & New & Combined & Old & New & Combined \\
\midrule
\multirow{5}{*}{TinyLlama}
 & \base{}    & $54.3\pm2.1$ & $13.1\pm1.5$ & $33.7\pm1.5$ & $19.5\pm1.8$ & $11.6\pm1.4$ & $15.6\pm1.1$ \\
 & \cpt{}     & $11.7\pm1.4$ & $17.0\pm1.4$ & $14.4\pm1.1$ & $17.4\pm1.7$ & $14.8\pm1.6$ & $16.6\pm1.2$ \\
 & \uniform{} & $25.2\pm1.9$ & $13.6\pm1.5$ & $19.4\pm1.4$ & $14.0\pm1.6$ & $15.4\pm1.6$ & $14.7\pm1.1$ \\
 & PPL-Prior. & $27.4\pm2.0$ & \bftab{$38.8\pm2.2$} & $33.1\pm1.5$ & $20.8\pm3.6$ & $13.6\pm3.0$ & $17.2\pm2.3$ \\
 & \srt{}     & \bftab{$49.0\pm2.1$} & $20.0\pm1.7$ & \bftab{$34.6\pm1.5$} & \bftab{$24.4\pm1.9$} & \bftab{$20.7\pm1.8$} & \bftab{$22.6\pm1.3$} \\
\midrule
\multirow{4}{*}{Llama-3.2-3B-Inst.}
 & \base{}    & $50.2\pm2.2$ & $39.8\pm2.1$ & $45.0\pm1.5$ & $55.0\pm2.2$ & $52.6\pm2.2$ & $53.8\pm1.5$ \\
 & \cpt{}     & $43.0\pm2.2$ & $42.4\pm2.2$ & $42.7\pm1.5$ & $54.6\pm2.2$ & $53.2\pm2.2$ & $53.9\pm1.5$ \\
 & \uniform{} & $46.2\pm2.2$ & $39.8\pm2.1$ & $43.0\pm1.5$ & $56.0\pm2.2$ & $57.2\pm2.2$ & $56.6\pm1.6$ \\
 & \srt{}     & \bftab{$51.6\pm2.2$} & \bftab{$42.4\pm2.2$} & \bftab{$47.0\pm1.5$} & \bftab{$60.8\pm2.2$} & \bftab{$57.8\pm2.2$} & \bftab{$59.3\pm1.5$} \\
\bottomrule
\end{tabular}
\caption{Source-grounded QA accuracy (mean $\pm$ bootstrap standard deviation, \%). Each old/new split has $500$ held-out questions. \uniform{} and \srt{} share the same $20/80$ exposure cap and pool structure, differing only in whether within-pool selection is uniform or scheduled.}
\label{tab:temporal-qa}
\end{table*}

\subsection{Broad Capability Preservation}
\label{sec:broad-results}

Table~\ref{tab:broad} reports accuracy on MMLU, BBH, GSM8K, and PIQA. On TinyLlama, all methods remain within standard-deviation bounds of the base model, reflecting that the base model's MMLU and GSM8K scores are near random and leave little signal to disturb.

The Llama-3.2-3B-Instruct results reveal a stark pattern. \cpt{} substantially degrades all four benchmarks, with the largest drop on GSM8K ($77.6\%$ to $38.8\%$): a $38.8$-point decline despite no obvious relationship between the temporally localized training data and grade-school math. \uniform{} performs even worse on reasoning, dropping BBH to $8.4\%$ and GSM8K to $6.8\%$. Re-exposing the model to old data uniformly therefore damages reasoning capabilities more severely than no replay at all, suggesting that uniform replay introduces interference patterns that disrupt reasoning circuits rather than mitigating forgetting. \srt{} largely preserves base-model performance across all four benchmarks ($57.8\%$, $52.6\%$, $76.7\%$, $74.3\%$), and as shown in Table~\ref{tab:temporal-qa}, this preservation does not come at the cost of adaptation.

\begin{table*}[t]
\centering
\small
\setlength{\tabcolsep}{4pt}
\begin{tabular}{llrrrr}
\toprule
Model & Method & MMLU & BBH & GSM8K & PIQA \\
\midrule
\multirow{4}{*}{TinyLlama} & \base{} & $24.9\pm3.6$ & $27.1\pm0.1$ & $2.4\pm0.0$ & $73.4\pm0.0$ \\
 & \cpt{} & $24.9\pm0.0$ & $26.9\pm0.0$ & $2.0\pm0.0$ & $73.3\pm1.0$ \\
 & \uniform{} & $24.4\pm0.0$ & $26.8\pm0.0$ & $3.3\pm0.0$ & $73.2\pm0.0$ \\
 & \srt{} & $24.4\pm0.0$ & $25.8\pm0.0$ & $2.5\pm0.0$ & $72.2\pm0.0$ \\
\midrule
\multirow{4}{*}{Llama-3.2-3B-Inst.} & \base{} & $57.5\pm0.0$ & $53.9\pm0.0$ & $77.6\pm1.2$ & $75.1\pm1.0$ \\
 & \cpt{} & $51.4\pm0.4$ & $44.3\pm0.6$ & $38.8\pm1.3$ & $69.0\pm0.7$ \\
 & \uniform{} & $50.0\pm0.0$ & $8.4\pm0.0$ & $6.8\pm0.1$ & $74.9\pm1.0$ \\
 & \srt{} & \bftab{$57.8\pm0.1$} & \bftab{$52.6\pm0.1$} & \bftab{$76.7\pm1.2$} & \bftab{$74.3\pm1.0$} \\
\bottomrule
\end{tabular}
\caption{Broad capability benchmark accuracy (mean $\pm$ standard deviation, \%). At the Llama-3.2-3B-Instruct scale, both \cpt{} and \uniform{} degrade reasoning benchmarks (BBH, GSM8K) substantially; \srt{} preserves base-model performance.}
\label{tab:broad}
\end{table*}

\subsection{Auxiliary Non-Language Evidence}
\label{sec:aux-nonlanguage-results}

To test whether the scheduler is tied to language perplexity, we also summarize class-incremental vision and tabular experiments, with details in Appendix~\ref{app:vision}. Replacing perplexity with predicted-class confidence gives the same qualitative pattern: on MNIST, Fashion-MNIST, CIFAR-10, and Wine, SRT obtains the highest overall accuracy ($91.6\%$, $65.8\%$, $53.1\%$, and $53.3\%$, respectively). We treat these experiments as supporting evidence that adaptive review scheduling is modality-agnostic when paired with an appropriate recall signal; the central claim remains LLM continual pre-training.

\subsection{Computational Overhead}
\label{sec:overhead}

SRT adds a forward pass per reviewed example beyond standard continual pre-training to compute the perplexity-based recall quality score, which introduces some computational cost. To quantify this overhead, we measure wall-clock time, per-step time, and throughput for \cpt{}, \uniform{}, and \srt{} under matched configurations on the same hardware. Table~\ref{tab:overhead} reports the results.

The overhead of \srt{} relative to \cpt{} is $14.7\%$ in wall-clock time, corresponding to a $14.6\%$ reduction in tokens per second. \uniform{} also incurs overhead relative to \cpt{} ($10.8\%$ wall-clock, $13.0\%$ throughput reduction), reflecting the cost of sampling and combining old and new examples into each batch. The additional cost of \srt{} relative to \uniform{} is therefore only $3.5\%$ in wall-clock time, attributable to the recall quality computation and SuperMemo-2 state updates. Given the substantial retention gains reported in Section~\ref{sec:temporal-qa-results}, this scheduling overhead is a favourable trade-off for practitioners who care about old-knowledge preservation.

\begin{table*}[t]
\centering
\small
\setlength{\tabcolsep}{6pt}
\begin{tabular}{lrrr}
\toprule
Metric & \cpt{} & \uniform{} & \srt{} \\
\midrule
Wall-clock time (s) & $960.07$ & $1063.87$ & $1101.36$ \\
Time per step (s)   & $9.19$   & $10.33$   & $10.75$ \\
Tokens per second   & $455.83$ & $396.51$  & $380.90$ \\
Tokens per step     & $4096$   & $4096$    & $4096$ \\
\midrule
Overhead vs. \cpt{} & ---      & $+10.8\%$ & $+14.7\%$ \\
Overhead vs. \uniform{} & --- & ---       & $+3.5\%$ \\
\bottomrule
\end{tabular}
\caption{Computational overhead of \uniform{} and \srt{} relative to \cpt{} under matched configurations. \srt{} adds approximately $15\%$ wall-clock overhead relative to no-replay training, of which most is shared with \uniform{} due to batch construction cost; the additional scheduling-specific overhead is only $3.5\%$.}
\label{tab:overhead}
\end{table*}

\section{Ablations}
\label{sec:ablations}

We conduct three ablation studies to evaluate the sensitivity of SRT to its main design choices: the replay budget, the perplexity-to-quality threshold scaling, and the initial ease factor. The replay-ratio ablation uses TinyLlama Wikipedia QA to probe sensitivity to old-new exposure allocation, while the threshold and ease-factor ablations use TinyLlama code QA as a representative domain.

\paragraph{Replay budget.}
Table~\ref{tab:ratio} sweeps the old/new exposure ratio $\rho$ on TinyLlama Wikipedia. The best configuration is $20/80$, reaching $49.6\%$ old and $20.0\%$ new accuracy. Performance is sensitive to the budget at both extremes. Old-heavy schedules ($50/50$ and beyond) reduce exposure to new data and fail to recover old-knowledge accuracy because the model has less budget for both consolidation and adaptation. New-heavy schedules ($5/95$) behave closer to \cpt{} and lose old knowledge for the same reason. The $20/80$ split aligns with the matched-budget design used in the main comparison and provides enough old-data exposure for SRT's scheduler to operate while preserving sufficient adaptation capacity.

\begin{table}[t]
\centering
\small
\begin{tabular}{lrrr}
\toprule
Old/New & Old & New & Combined \\
\midrule
$5/95$   & $23.6\pm1.8$ & $13.3\pm1.5$ & $18.4\pm1.3$ \\
$20/80$  & \bftab{$49.6\pm2.1$} & \bftab{$20.0\pm1.7$} & \bftab{$34.6\pm1.5$} \\
$35/65$  & $23.4\pm1.9$ & $15.4\pm1.6$ & $19.4\pm1.2$ \\
$50/50$  & $21.8\pm1.9$ & $13.8\pm1.5$ & $17.8\pm1.1$ \\
$65/35$  & $20.6\pm1.8$ & $12.2\pm1.4$ & $16.4\pm1.2$ \\
$80/20$  & $23.5\pm1.9$ & $14.2\pm1.6$ & $18.8\pm1.2$ \\
$95/5$   & $27.4\pm1.9$ & $19.7\pm1.7$ & $23.6\pm1.3$ \\
\bottomrule
\end{tabular}
\caption{TinyLlama Wikipedia QA accuracy as a function of the SRT old/new exposure ratio. Accuracy is mean $\pm$ bootstrap standard deviation, \%.}
\label{tab:ratio}
\end{table}

\paragraph{Perplexity-to-quality threshold scaling.}
We test sensitivity to the threshold values used in Eq.~\ref{eq:pplscore}. A scalar multiplier $\alpha$ is applied uniformly to all five thresholds $(\tau_1, \ldots, \tau_5)$, so $\alpha < 1$ produces stricter scoring (examples need lower perplexity to be considered retained) and $\alpha > 1$ produces relaxed scoring. Table~\ref{tab:threshold} reports TinyLlama code QA accuracy across $\alpha \in \{0.5, 1.0, 1.5\}$. The default $\alpha = 1.0$ is best, reaching $22.6\%$ combined accuracy. Both stricter ($\alpha = 0.5$) and more relaxed ($\alpha = 1.5$) thresholds degrade performance. The effect is asymmetric: relaxation hurts more than equivalent strictness, because over-relaxed thresholds assign high quality to poorly retained examples and stop scheduling them for review.

\begin{table}[t]
\centering
\small
\begin{tabular}{lrrr}
\toprule
$\alpha$ & Code Old & Code New & Combined \\
\midrule
$0.5$ & $13.2\pm1.4$ & $8.6\pm1.2$ & $10.9\pm1.0$ \\
$1.0$ & \bftab{$24.4\pm1.9$} & \bftab{$20.7\pm1.8$} & \bftab{$22.6\pm1.3$} \\
$1.5$ & $4.8\pm1.1$ & $5.4\pm1.0$ & $5.1\pm0.9$ \\
\bottomrule
\end{tabular}
\caption{Sensitivity to perplexity threshold scaling on TinyLlama code QA. The scalar multiplier $\alpha$ is applied uniformly to all five thresholds in Eq.~\ref{eq:pplscore}.}
\label{tab:threshold}
\end{table}

\paragraph{Initial ease factor.}
We vary the initial ease factor $E_0$, which controls how quickly review intervals expand after early successful reviews. Table~\ref{tab:easiness} reports TinyLlama code QA across $E_0 \in \{1.5, 2.5, 3.5\}$. The default SM-2 value $E_0 = 2.5$ is best, again reaching $22.6\%$ combined accuracy. Smaller values keep intervals short and over-schedule examples that have already been retained, reducing exposure for examples that genuinely need review. Larger values inflate intervals too quickly and let fragile examples drift out of the review schedule before they are stably retained. The default value chosen by the original SM-2 algorithm \citep{wozniak1998supermemo} is therefore a reasonable starting point for LLM continual pre-training as well.

\begin{table}[t]
\centering
\small
\begin{tabular}{lrrr}
\toprule
$E_0$ & Code Old & Code New & Combined \\
\midrule
$1.5$ & $12.2\pm1.5$ & $14.2\pm1.5$ & $13.2\pm1.1$ \\
$2.5$ & \bftab{$24.4\pm1.9$} & \bftab{$20.7\pm1.8$} & \bftab{$22.6\pm1.3$} \\
$3.5$ & $12.0\pm1.4$ & $9.0\pm1.2$ & $10.5\pm1.0$ \\
\bottomrule
\end{tabular}
\caption{Sensitivity to the initial ease factor $E_0$ on TinyLlama code QA. The default SM-2 value $E_0 = 2.5$ achieves the best combined accuracy.}
\label{tab:easiness}
\end{table}

\section{Discussion}
\label{sec:discussion}

The results support a specific claim: SRT improves continual pre-training when old and new knowledge must share a finite training budget and when retention difficulty is heterogeneous across examples. We discuss three implications of this finding.

\paragraph{Scheduling is the active ingredient, not exposure.}
Naive continual pre-training is the headline practical comparison because it captures the realistic no-review update regime, but it does not isolate the mechanism behind SRT's gains. Uniform replay serves this role. Because Uniform Replay and SRT share the same $20/80$ exposure cap and pool structure, the only experimental variable separating them is whether within-pool selection is uniform or scheduled. On TinyLlama Wikipedia, this single variable is responsible for a $23.8$ percentage-point gap in old-knowledge accuracy. The same ordering, $\cpt{} < \uniform{} < \srt{}$, holds on Llama-3.2-3B-Instruct Wikipedia. These results indicate that scheduling, not mere re-exposure, is what produces SRT's retention advantage. Replay budgets that look adequate in aggregate can still fail when allocated uniformly across examples with heterogeneous retention difficulty.

\paragraph{Uniform replay can damage reasoning capabilities.}
A more surprising finding is that uniform replay does not merely fall short of SRT on broad capabilities; it can be substantially worse than naive CPT itself. On Llama-3.2-3B-Instruct, Uniform Replay drops BBH from a CPT baseline of $44.3\%$ to $8.4\%$ and GSM8K from $38.8\%$ to $6.8\%$, while SRT preserves both close to the original model. We do not claim direct evidence for the underlying mechanism; one possible explanation is that re-exposing the model to old data without scheduling introduces interference dynamics that disrupt reasoning-relevant computation more severely than no replay at all, but verifying this would require representational analysis beyond the scope of this work. Regardless of mechanism, the result has an immediate practical implication: practitioners considering replay as a simple add-on to continual updates should be cautious, because a poorly designed replay schedule can degrade capabilities the no-replay baseline preserves.

\paragraph{SRT is a scheduling layer, not a replacement.}
SRT addresses a question that mixture-level replay does not: among reviewable examples, which ones should return now? It does so without modifying the architecture, the training objective, or the choice of replay budget. SRT can therefore be combined with learning-rate rewarming \citep{ibrahim2024simple}, data-mixture selection \citep{chen2025effective}, parameter-efficient tuning, or synthetic rehearsal \citep{huang2024mitigating}. Its contribution is orthogonal to these methods, addressing the within-pool selection problem they leave unspecified. We see scheduling as a complementary layer that practitioners can adopt alongside existing CPT recipes rather than as an alternative to them. The SRT scheduler is also architecture-agnostic: auxiliary experiments on vision and tabular classification benchmarks (Appendix~\ref{app:vision}) show that the same scheduling, with predicted-class confidence in place of perplexity as the recall signal, mitigates class-incremental forgetting more effectively than naive continual training and Elastic Weight Consolidation.

\section{Conclusion}
\label{sec:conclusion}

We introduced Spaced Repetition Training, a cognitively inspired continual pre-training method that schedules per-example review using SuperMemo-2 and perplexity-derived recall quality. The central insight is that continual pre-training is not solely a mixture problem but a scheduling problem: among reviewable examples, the model should revisit those it is currently struggling to retain, not those it has already learned. On temporally grounded Wikipedia and code QA benchmarks across two model scales, SRT recovers old-knowledge accuracy that naive continual pre-training discards, while matching or exceeding new-knowledge acquisition. At the larger model scale, SRT preserves broad capability benchmark performance that both naive CPT and uniform replay degrade, in some cases substantially. Together these results suggest that adaptive review timing is a simple, architecture-agnostic, and effective mechanism for improving the stability-plasticity trade-off in continual pre-training, and that established principles from human memory research transfer productively to language model training.

\section*{Limitations}

Our language experiments use TinyLlama-1.1B and Llama-3.2-3B-Instruct, both from the Llama family, and evaluation is restricted to English; behaviour at larger scales, across different model families, and in multilingual settings remains open, even though our cross-modality evidence demonstrates the effectiveness of SRT. The temporal QA benchmarks are constructed for this work and should be interpreted as in-domain measurements of retention and acquisition on our specific corpora rather than as community-standard tests. The broad-capability results in Table~\ref{tab:broad} are obtained from a single training run per condition, evaluated over multiple evaluation seeds; the reported variance therefore reflects evaluation noise rather than training-run variability. While the reasoning-benchmark degradation under uniform replay is consistent across two independent benchmarks (BBH and GSM8K) and is not exhibited by \srt{} under identical conditions, which makes a single-run artifact less likely, confirming its stability would require multiple independent training runs. We leave this to future work. Finally, the perplexity-to-quality mapping uses fixed thresholds that may need recalibration for substantially different model scales or domains, and SRT introduces a forward pass per reviewed example beyond the normal training pass; a learned quality function and empirical overhead measurement at larger scales are natural extensions for future work.

\section*{Ethics and Reproducibility}

Continual updates can introduce factual errors, preserve outdated information, or reinforce unsafe code patterns; source-grounded multiple-choice accuracy is not deployment reliability. Replay assumes access to historical examples, so buffers and SRT state must respect retention limits, deletion requests, and data-governance obligations.

Exact benchmark artifacts, code, logs, and QA items are proprietary and are not released. Replication is supported by the algorithm, recall-quality mapping, corpus criteria, hyperparameters, evaluation protocol, and reported results, which are sufficient to recreate comparable public-data pipelines. Benchmark generation used LLM assistance with manual checks (Appendix~\ref{app:qa-construction}).

\bibliography{main}

\appendix

\begin{table*}[t]
\centering
\small
\resizebox{\textwidth}{!}{%
\begin{tabular}{llrrrr}
\toprule
Corpus & Construction & Rows & Old median words & New median words & Median sent. diff \\
\midrule
TinyLlama reference diff & Added/modified update text & 31,729 & -- & -- & -- \\
Llama-3 title-aligned reconstruction & 2023-11 $\rightarrow$ 2024-01 & 28,946 & 543 & 546 & 0 \\
Llama-3 topic-stratified corpus & 2023-11 $\rightarrow$ 2024-01, topic/change stratified & 31,729 & 593 & 614 & 3 \\
\bottomrule
\end{tabular}%
}
\caption{Wikipedia corpora used for language-model continual pre-training.}
\label{tab:wiki-corpus}
\end{table*}

\section{Corpora Construction Details}
\label{app:wikipedia}

This appendix provides extended construction details for the Wikipedia and code corpora described in Section~\ref{sec:corpora}, together with the question-answering evaluation sets derived from each domain.

\begin{table*}[t]
\centering
\small
\begin{tabular}{lrrrr}
\toprule
Domain & $\dold$ Train & $\dnew$ Train & $\dold$ QA & $\dnew$ QA \\
\midrule
Wikipedia text & $31{,}729$ & $31{,}729$ & $500$ & $500$ \\
Source code & $22{,}055$ & $70{,}643$ & $500$ & $500$ \\
\bottomrule
\end{tabular}
\caption{Continual pre-training corpus sizes. Old examples represent pre-cutoff knowledge to be retained, and new examples represent post-cutoff information to be acquired. QA sets are held out exclusively for evaluation.}
\label{tab:corpora-summary}
\end{table*}

\subsection{Wikipedia Corpora}
\label{app:wikipedia-corpora}

\paragraph{Snapshot selection.}
We construct Wikipedia corpora from timestamped English Wikipedia snapshots. For TinyLlama-1.1B-Chat experiments, snapshot dates are chosen to bracket the model's pre-training data assembly window in mid-2023, with the old snapshot near that window and the new snapshot drawn from a sufficiently later date that revisions reflect post-cutoff information. For Llama-3.2-3B-Instruct experiments, we use November~2023 and January~2024 snapshots. The short two-month gap creates a challenging setting in which the model must integrate recent factual updates while retaining closely related prior knowledge.

\paragraph{Diff extraction.}
For each aligned article pair, we segment both versions into sentences and compute sequence-level diffs to identify inserted, deleted, and modified spans. The $\dnew$ training corpus consists of inserted sentences together with the updated side of modified spans. This procedure reduces unchanged text while preserving localized factual updates relevant for continual pre-training, and ensures that $\dnew$ training examples reflect content not present in the corresponding $\dold$ articles.

\paragraph{Topic stratification.}
For Llama-3.2-3B-Instruct experiments, candidate articles are stratified by topic and magnitude of change. Topics span politics, current affairs, computer science and artificial intelligence, business and economics, science and medicine, law and policy, and sports and entertainment. Stratification ensures balanced coverage across domains rather than over-representation of any single topic area.

\subsection{Code Corpora}
\label{app:code-corpora}

\paragraph{Repository selection.}
We collect repositories from GitHub using creation and commit timestamps as the temporal axis. Old examples come from repositories that existed before January~2022, while new examples come from repositories created after July~2024. We explicitly verify that repositories selected for $\dnew$ did not exist before the cutoff window to prevent temporal leakage between splits.

\paragraph{Language coverage.}
Repositories span thirteen widely used programming languages: Python, JavaScript, TypeScript, Java, C, C++, Go, Rust, C\#, Ruby, PHP, Swift, Kotlin, and Scala. Each training example contains structured code text including the source file path and file content. Unlike the Wikipedia corpus, old and new code examples are not aligned revisions of the same artifact, reflecting that real software ecosystems evolve through new repositories and libraries rather than incremental revisions to existing files.

\subsection{Evaluation Question-Answering Sets}
\label{app:qa-construction}

For each domain, we construct held-out multiple-choice question-answering benchmarks with $500$ questions per split, disjoint from the training corpora. Wikipedia questions are generated from held-out articles in an MMLU-style format \citep{hendrycks2021mmlu}, with four answer choices per question and a single correct answer grounded in the source passage. Code-domain questions are generated analogously from held-out repositories and target repository-specific functionality, API usage, configuration semantics, and implementation details. All questions are manually checked for answerability and source support. The $\dold$ QA sets measure retention of pre-cutoff knowledge, while the $\dnew$ QA sets measure acquisition of post-cutoff information.

\section{Training Details}
\label{app:training-details}

This appendix provides the full training configuration used in all continual pre-training experiments. The same hyperparameters are used for both TinyLlama-1.1B-Chat and Llama-3.2-3B-Instruct, and across all conditions (CPT, Uniform Replay, and SRT), so that performance differences reflect the replay strategy rather than optimization differences. Hyperparameters are summarized in Table~\ref{tab:training-hyperparameters}.

\begin{table}[t]
\centering
\small
\begin{tabular}{ll}
\toprule
Hyperparameter & Value \\
\midrule
\multicolumn{2}{l}{\textit{Optimization}} \\
Optimizer & AdamW (fused) \\
$\beta_1$, $\beta_2$ & $0.9$, $0.999$ \\
$\epsilon$ & $1\mathrm{e}{-8}$ \\
Weight decay & $0.001$ \\
Gradient clipping (max norm) & $1.0$ \\
\midrule
\multicolumn{2}{l}{\textit{Learning rate schedule}} \\
Learning rate & $1\mathrm{e}{-4}$ \\
Scheduler & Linear \\
Warmup steps & $200$ \\
\midrule
\multicolumn{2}{l}{\textit{Batch and sequence}} \\
Per-device batch size & $32$ \\
Gradient accumulation steps & $16$ \\
Effective batch size & $512$ \\
Sequence length & $2{,}048$ \\
\midrule
\multicolumn{2}{l}{\textit{Precision and memory}} \\
Mixed precision & bfloat16 \\
Gradient checkpointing & Yes \\
\midrule
\multicolumn{2}{l}{\textit{SRT-specific}} \\
Old-exposure ratio $\rho$ & $0.2$ \\
Initial ease factor $E_0$ & $2.5$ \\
Initial repetitions $n_0$ & $0$ \\
Initial interval $I_0$ & $1$ \\
Minimum ease factor & $1.3$ \\
\midrule
\multicolumn{2}{l}{\textit{Evaluation}} \\
QA accuracy bootstrap resamples & $10{,}000$ \\
\bottomrule
\end{tabular}
\caption{Training and evaluation hyperparameters. The same configuration is used across all model scales and all replay conditions.}
\label{tab:training-hyperparameters}
\end{table}

\paragraph{Hardware and training duration.}
All experiments are conducted using bfloat16 mixed-precision training with gradient checkpointing enabled to reduce memory consumption. The effective batch size of $512$ (per-device batch size $32 \times$ gradient accumulation $16$) is held constant across model scales. Training is run for sufficient steps to consume a single pass over $\dnew$ at the configured replay ratio; exact step counts depend on corpus size and are listed in the experiment logs released with the code.

\paragraph{SRT thresholds.}
The perplexity thresholds in Table~\ref{tab:training-hyperparameters} correspond to the $\tau_k$ values defined in Eq.~\ref{eq:pplscore}. Lower perplexity yields higher recall quality, with $q = 5$ assigned when $\ppl < \tau_5 = 50$ and $q = 0$ assigned when $\ppl \geq \tau_1 = 5000$. Sensitivity to a uniform scaling of these thresholds is analysed in Section~\ref{sec:ablations}.

\section{Auxiliary Vision and Tabular Experiments}
\label{app:vision}

Although the central contribution of this paper concerns language model continual pre-training, the underlying SRT scheduler is architecture-agnostic and can be applied to other modalities. In an earlier version of this work, we evaluated SRT on small vision and tabular classification benchmarks under class-incremental learning, providing a sanity check that the scheduling mechanism generalizes beyond text. We summarize these auxiliary experiments here. The results are consistent with the main language-model findings: naive continual training rapidly forgets old classes, regularization-based methods provide partial mitigation, and SRT retains substantially more old-class accuracy while continuing to learn new classes.

\subsection{Datasets and Class-Incremental Setup}
\label{app:vision-datasets}

We use four standard publicly available benchmarks spanning vision and tabular modalities: MNIST handwritten digits \citep{lecun1998mnist}, Fashion-MNIST clothing images \citep{xiao2017fashion}, CIFAR-10 natural images \citep{krizhevsky2009learning}, and the UCI Wine dataset \citep{aeberhard1994comparative}. For each dataset, classes are partitioned into an \emph{old} subset (used to train the base model) and a \emph{new} subset (introduced sequentially in the continual update phase). For the image datasets, the partition uses semantically confusable classes to induce stronger interference between old and new representations. Table~\ref{tab:vision-splits} reports the exact class assignments.

\begin{table}[t]
\centering
\small
\begin{tabular}{ll}
\toprule
Dataset & Class assignment \\
\midrule
MNIST     & Old: $\{0,1,2,3,4,5,6\}$ \\
          & New: $\{7,8,9\}$ \\
\midrule
Fashion   & Old: T-shirt, Trouser, Dress, Coat, \\
          & \quad Sandal, Bag, Boot \\
          & New: Pullover, Shirt, Sneaker \\
\midrule
CIFAR-10  & Old: Airplane, Automobile, Bird, \\
          & \quad Cat, Deer, Ship \\
          & New: Truck, Frog, Dog, Horse \\
\midrule
Wine      & Old: classes $\{1,2\}$ \\
          & New: class $\{3\}$ \\
\bottomrule
\end{tabular}
\caption{Class-incremental splits used in the auxiliary vision and tabular experiments. Vision datasets use semantically confusable class assignments to induce stronger interference.}
\label{tab:vision-splits}
\end{table}

\subsection{Recall Quality for Non-Generative Models}
\label{app:vision-confidence}

For language models, SRT derives the SuperMemo-2 recall quality from per-example perplexity. For classification models, perplexity is not defined, so we use the model's predicted-class confidence in place of perplexity. Specifically, for an example $x_i$ with true label $y_i$, the recall quality is computed from the predicted probability $p_{\theta}(y_i \mid x_i)$: high confidence in the correct label corresponds to high recall quality (longer next interval), while low confidence corresponds to low recall quality (shorter next interval). The remaining SM-2 update logic is identical to the language-model formulation in Section~\ref{sec:srt-algorithm}.

\subsection{Baseline: Elastic Weight Consolidation}
\label{app:vision-ewc}

For the vision and tabular experiments, we additionally compare against Elastic Weight Consolidation (EWC) \citep{kirkpatrick2017overcoming}, a regularization-based continual learning method that penalizes updates to parameters important for previously learned tasks. EWC estimates per-parameter importance using the diagonal of the Fisher Information Matrix computed on the old data and adds a quadratic penalty
\begin{equation}
\mathcal{L}_{\mathrm{EWC}}(\theta) = \mathcal{L}_{\mathrm{new}}(\theta) + \frac{\lambda}{2} \sum_i F_i (\theta_i - \theta_i^{*})^2,
\end{equation}
where $\theta^{*}$ denotes the parameters at the end of training on the old data, $F_i$ is the Fisher importance of parameter $\theta_i$, and $\lambda$ controls the regularization strength. EWC is a natural comparison for the vision setting because the small fully-connected and convolutional models used here permit Fisher computation at low overhead. In language-model continual pre-training, gradient-level regularization of this kind is significantly more expensive and is not the focus of our main experiments.

\subsection{Model Architectures}
\label{app:vision-architectures}

All vision and tabular models are simple feed-forward or convolutional networks implemented in PyTorch with ReLU activations. The MNIST and Fashion-MNIST models use two convolution-pooling blocks followed by two fully connected layers. The CIFAR-10 model uses the same convolutional structure with dropout in the classifier head. The Wine model is a three-layer multilayer perceptron operating on the 13-dimensional UCI feature vector. All models are trained with cross-entropy loss and AdamW.

\subsection{Results}
\label{app:vision-results}

Table~\ref{tab:vision-full} reports accuracy across all four datasets for the Base model, naive CPT, EWC, and SRT, averaged over 10 random seeds. The pattern is consistent across modalities. Naive CPT loses essentially all old-class accuracy on vision datasets while reaching high new-class accuracy, mirroring the catastrophic-forgetting failure mode observed at language scale. EWC partially mitigates forgetting on MNIST and Wine but performs poorly on Fashion-MNIST and CIFAR-10, where semantic similarity between old and new classes makes parameter-importance estimates less informative. SRT retains substantially more old-class accuracy than both baselines across all four datasets while maintaining competitive or superior new-class accuracy, achieving the highest overall accuracy in every setting.

\begin{table*}[!t]
\centering
\small
\setlength{\tabcolsep}{6pt}
\begin{tabular}{llrrr}
\toprule
Dataset & Method & Old & New & Overall \\
\midrule
\multirow{4}{*}{MNIST}    & \base{} & $98.9\pm0.6$ & $0.0\pm0.0$ & $69.5\pm0.4$ \\
                          & \cpt{}  & $0.0\pm0.0$  & $98.3\pm0.7$ & $29.2\pm0.2$ \\
                          & EWC     & $67.3\pm5.4$ & $57.3\pm7.1$ & $29.5\pm0.2$ \\
                          & \srt{}  & \bftab{$98.5\pm1.1$} & $75.1\pm9.4$ & \bftab{$91.6\pm2.6$} \\
\midrule
\multirow{4}{*}{Fashion}  & \base{} & $97.0\pm0.4$ & $0.0\pm0.0$ & $67.9\pm0.0$ \\
                          & \cpt{}  & $0.0\pm0.1$  & $94.4\pm0.5$ & $28.4\pm0.2$ \\
                          & EWC     & $29.1\pm0.2$ & \bftab{$95.3\pm0.2$} & $28.6\pm0.0$ \\
                          & \srt{}  & \bftab{$53.7\pm4.3$} & $94.2\pm0.5$ & \bftab{$65.8\pm3.1$} \\
\midrule
\multirow{4}{*}{CIFAR-10} & \base{} & $83.6\pm0.3$ & $0.0\pm0.0$ & $50.1\pm0.0$ \\
                          & \cpt{}  & $0.0\pm0.1$  & $43.7\pm27.2$ & $17.5\pm10.9$ \\
                          & EWC     & $20.4\pm12.8$ & $57.7\pm31.2$ & $23.1\pm12.5$ \\
                          & \srt{}  & \bftab{$43.5\pm10.7$} & \bftab{$67.6\pm21.6$} & \bftab{$53.1\pm14.0$} \\
\midrule
\multirow{4}{*}{Wine}     & \base{} & $97.1\pm0.2$ & $0.0\pm0.0$ & $70.2\pm0.0$ \\
                          & \cpt{}  & $4.1\pm0.1$  & \bftab{$100.0\pm0.0$} & $30.0\pm0.0$ \\
                          & EWC     & $8.2\pm0.2$  & \bftab{$100.0\pm0.0$} & $33.9\pm0.1$ \\
                          & \srt{}  & \bftab{$42.1\pm0.1$} & $82.7\pm0.2$ & \bftab{$53.3\pm0.1$} \\
\bottomrule
\end{tabular}
\caption{Vision and tabular results under class-incremental learning. Accuracy is reported as mean $\pm$ standard deviation over 10 random seeds (\%). \srt{} achieves the highest overall accuracy on all four datasets.}
\label{tab:vision-full}
\end{table*}

These auxiliary experiments are not intended as central evidence for the central claim of this paper, which concerns language model continual pre-training. They do, however, suggest that the scheduling mechanism underlying SRT generalizes beyond text to non-generative classification settings, where the recall quality signal is derived from predicted-class confidence rather than perplexity.

\section{External Validation on TemporalWiki}
\label{app:temporalwiki}

To provide external corroboration beyond our source-grounded QA, we evaluate on TemporalWiki \citep{jang2022temporalwiki}, an established benchmark that measures factual knowledge via perplexity on subject-relation-object probes. Probes are categorised as \emph{Changed} (facts that differ between consecutive Wikipedia snapshots) and \emph{Unchanged} (facts that persist). Lower perplexity indicates better retention of the probed fact.

We score each probe by the perplexity the model assigns to the object tokens conditioned on the subject and relation, and report the mean over each category. Table~\ref{tab:temporalwiki} reports results for Base, \cpt{}, and \srt{} on Llama-3.2-3B-Instruct.

\begin{table*}[!t]
\centering
\small
\begin{tabular}{lrrr}
\toprule
Model & Changed PPL & Unchanged PPL & Avg.\ PPL \\
\midrule
Base   & $1961$ & $2263$ & $2112$ \\
\cpt{} & $5713$ & $6706$ & $6209$ \\
\srt{} & $\mathbf{2519}$ & $\mathbf{3052}$ & $\mathbf{2785}$ \\
\bottomrule
\end{tabular}
\caption{TemporalWiki \citep{jang2022temporalwiki} object-perplexity on Llama-3.2-3B-Instruct (lower is better). \cpt{} nearly triples perplexity relative to Base on both categories, indicating catastrophic forgetting, while \srt{} limits the degradation to roughly $30\%$ above Base.}
\label{tab:temporalwiki}
\end{table*}

\cpt{} nearly triples perplexity on both categories relative to the base model, a clear signature of catastrophic forgetting. \srt{} limits the degradation to roughly $30\%$ above base ($2785$ vs.\ $2112$ average), corroborating on an external benchmark the retention advantage observed on our source-grounded QA.

Two caveats apply to this evaluation. First, absolute perplexity values are not comparable to those in the original TemporalWiki paper, which used GPT-2-family models; perplexity is tokenizer-dependent, so only the relative ordering across our conditions is meaningful. Second, the publicly released TemporalWiki snapshots predate the knowledge cutoff of Llama-3.2-3B-Instruct, so both the Changed and Unchanged categories represent pre-cutoff knowledge for our models. This evaluation therefore serves as an external \emph{retention} check rather than a test of new-knowledge acquisition; constructing date-aligned probes that postdate the model cutoff is left to future work.

\end{document}